\documentclass[12pt]{spieman}  
\usepackage{amsmath,amsfonts,amssymb}
\usepackage{graphicx}
\usepackage{setspace}
\usepackage{subfigure}
\usepackage[subfigure]{tocloft}
\usepackage{hyperref}
\usepackage{multirow}
\usepackage{booktabs,array}
\usepackage{makecell}

\usepackage{amssymb}
\usepackage{amsmath}
\usepackage{comment}
\usepackage[table]{xcolor}

\title{Relational Reasoning Network for Anatomical Landmarking}

\author[a]{Neslisah Torosdagli}
\author[a,f]{Syed Anwar}
\author[c]{Payal Verma}
\author[c]{Denise K. Liberton}
\author[c]{Janice S. Lee}
\author[d,e]{Wade W. Han}
\author[a,e,f]{Ulas Bagci}
\affil[a]{University of Central Florida, Orlando, FL}
\affil[b]{Sheikh Zyed Children's Hospital and George Washington University, Washington, DC}
\affil[c]{ Craniofacial Anomalies and Regeneration section, National Institute of Dental and Craniofacial Research (NIDCR), National Institutes of Health (NIH), Bethesda, MD.}
\affil[d]{Department of Otolaryngology - Head and Neck Surgery, Boston Children's Hospital, Harvard Medical School, Boston, MA.}
\affil[e]{Machine and Hybrid Intelligence Lab, Departments of Radiology, BME, and ECE, Northwestern University, Chicago, IL.}
\affil[f]{Ther-AI, LLC, Kissemmee, FL.}

\cftpagenumbersoff{figure}
\cftpagenumbersoff{table} 
\begin{document} 
\maketitle

\begin{abstract}

\noindent\textbf{Purpose:} We perform anatomical landmarking for craniomaxillofacial (CMF) bones without explicitly segmenting them. Towards this, we propose a new simple yet efficient deep network architecture, called \textit{relational reasoning network (RRN)}, to accurately learn the local and the global relations among the landmarks in CMF bones; specifically, mandible, maxilla, and nasal bones. \\
\textbf{Approach:} The proposed RRN works in an end-to-end manner, utilizing learned relations of the landmarks based on dense-block units. For a given few landmarks as input, RRN treats the landmarking process similar to a data imputation problem where predicted landmarks are considered missing.\\
\textbf{Results:}  We applied RRN to cone beam computed tomography scans obtained from 250 patients. With a 4-fold cross validation technique, we obtained an average root mean squared error of less than 2 mm per landmark. Our proposed RRN has revealed unique relationships among the landmarks that help us in inferring several \textit{reasoning} about informativeness of the landmark points. The proposed system identifies the missing landmark locations accurately even when severe pathology or deformation are present in the bones. \\ 
\textbf{Conclusions:} Accurately identifying anatomical landmarks is a crucial step in deformation analysis and surgical planning for CMF surgeries. Achieving this goal without the need for explicit bone segmentation addresses a major limitation of segmentation based approaches, where segmentation failure (as often the case in bones with severe pathology or deformation) could easily lead to incorrect landmarking. To the best of our knowledge, this is the first of its kind algorithm finding anatomical relations of the objects using deep learning.
\end{abstract}


{\noindent \footnotesize\textbf{*}Ulas Bagci,  \linkable{ulasbagci@gmail.com} }

\begin{spacing}{2}   

\section{Introduction}
\label{sect:intro}  
\begin{figure*}[t]
    \includegraphics[width=0.5\columnwidth]{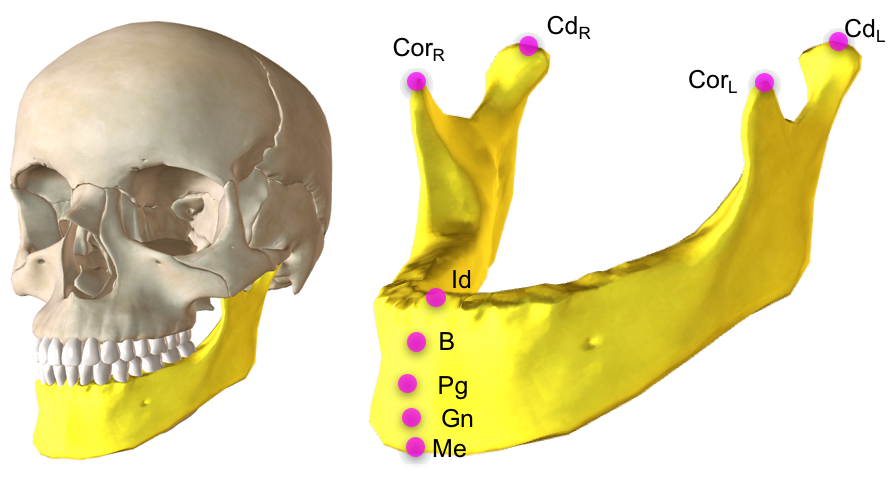}
    \includegraphics[width=0.5\columnwidth]{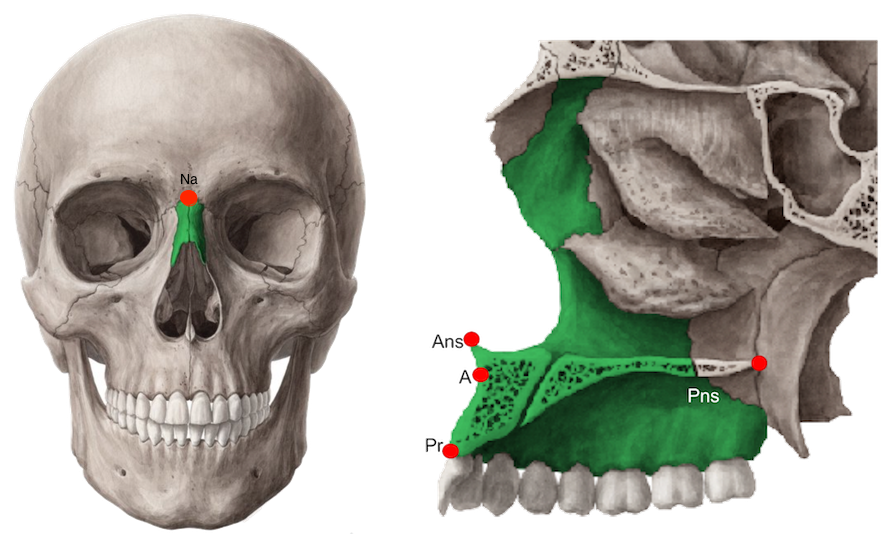}
    \caption{Mandible and Maxilla/Nasal bone anatomies, a) Mandibular Landmarks: Menton $(Me)$, Condylar Left $(Cd_L)$, Condylar Right $(Cd_R)$ , Coronoid Left $(Cor_L)$, Coronoid Right $(Cor_R)$, Infradentale$(Id)$, B point $(B)$, Pogonion $(Pg)$, and Gnathion $(Gn)$, b) Maxillary Landmarks: Anterior Nasal Spine $(ANS)$, Posterior Nasal Spine $(PNS)$,  A-Point $(A)$, and Prostion $(Pr)$, Nasal Bones Landmark: Nasion $(Na)$.}
    \label{fig:madible_maxilla}
\end{figure*}

In the United States alone, more than $17$ million patients suffer from developmental deformities of the jaw, face, and skull region due to trauma, deformities from tumor ablation, or congenital birth defects~\cite{Xia2009}. The number of patients who require orthodontic treatment is far beyond this number. An accurate anatomical landmarking on medical scans (mostly it is volumetric computed tomography-CT-scans) is a crucial step in the deformation analysis and surgical planning of the craniomaxillofacial (CMF) bones. This, if done correctly and efficiently, would afford precise image-based surgical planning for patients. This is even more significant since such deformities are known to vary from patient to patient and hence need careful delineation. However, manual landmarking in volumetric CT scans is a tedious process and prone to inter-operator variability. There are considerable efforts towards developing a fully-automated and accurate software for anatomical landmarking based on bone segmentation from CT scans~\cite{zhang_2017, automated_landmarking_challanges,lang2020automatic}. Despite this clinical need, very little progress has been made especially for bones with a high level of congenital and developmental deformations (approximately 5$\%$ of the CMF deformities).
\begin{figure}
    \includegraphics[width=0.5\columnwidth]{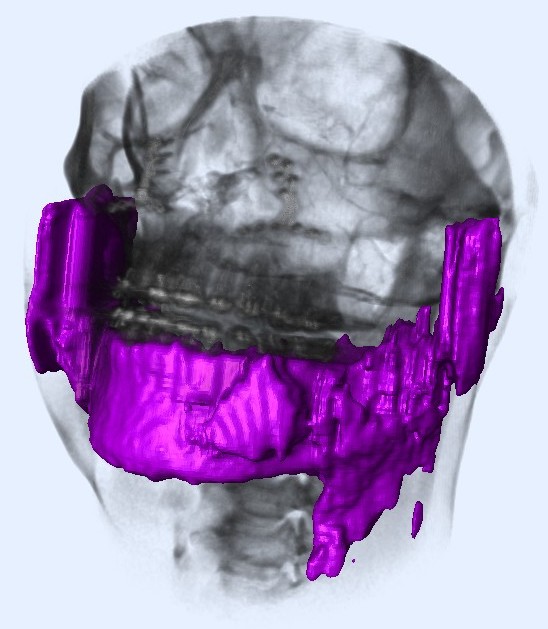}
    \includegraphics[width=0.5\columnwidth]{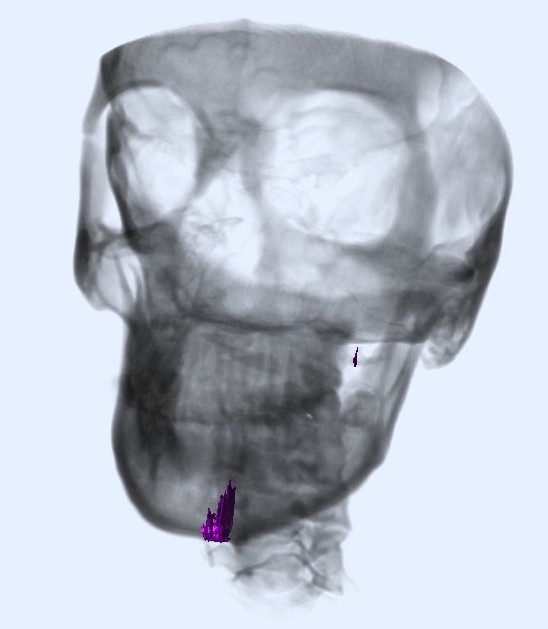}
    \caption{CT segmentation results rendered in fuchsia  which are scored as ``\textit{unacceptable segmentation}'' at~\cite{ness}, a) Patient with surgical intervention, b) Patient with high variability in the bone.}
    \label{fig:failing}
\end{figure}
Deep learning based approaches have become the standard choice for pixel-wise medical-image segmentation applications due to their high efficacy~\cite{zhang_2017, ness, ourMandibleSegmentation}. However, it is difficult to generalize segmentation especially when there is a high degree of deformation or pathology \cite{zhang2020context}, which is the case for treating CMF conditions. Figure~\ref{fig:failing} demonstrates two examples of challenging mandible cases where the patients have surgical intervention (left) and high variability in the bone (right), and causing segmentation algorithms to fail (leakage or under-segmentation). Current state-of-the-art landmarking algorithms are mostly dependent on bone segmentation results, since locating landmarks could become easier once their parent anatomy (the bones they belong to) is  precisely known \cite{zhang2020context}. Figure~\ref{fig:madible_maxilla} demonstrates mandible and maxilla/nasal bone anatomies along with the landmarks associated with those bones. If the underlying segmentation is poor, it is highly likely to have a large landmark localization error, directly affecting the quantification process (which could include severity measurement, surgical modeling, and treatment planing). 

We hypothesize that if an explicit segmentation can be avoided for extremely challenging cases, landmark localization errors can be minimized. This will also lead to a widespread use of landmarking procedure in surgical planning and precision medicine. Since, CMF bones are found in the same anatomical space even when there is deformity or pathology. Hence, the overall global relationship of the anatomical landmarks should still be preserved despite severe localized changes. Based on this rationale, we claim that utilizing local and global relations of the landmarks can help automatic landmarking without the extreme need for segmentation. 

\subsection{Background and Related Work}
\subsubsection{Landmarking}
Anatomical landmark localization approaches can broadly be categorized into three main groups: registration-based (atlas-based) \cite{Shahidi2014}, knowledge-based \cite{negrillo2020automatic, automated_landmarking_challanges_1}, and learning-based~\cite{dinggang_2017_survey, zhang2020context}. Integration of shape and appearance increases the accuracy of the registration-based approaches. However, image registration is still an ill-posed problem, and when there are variations such as age (pediatrics vs. adults), missing teeth (very common in certain age groups), missing bone or bone parts, severe pathology (congenital or trauma), and imaging artifacts, the performance can be quite poor~\cite{automated_landmarking_challanges, registration_landmarking, dir_cons}. The same concerns apply to segmentation based approaches too.

Gupta et al.~\cite{automated_landmarking_challanges_1} developed a knowledge-based algorithm to identify 20 anatomical landmarks on cone-beam CT (CBCT) scans. Despite their promising results, a seed must be selected by using $3D$ template registration on the inferior-anterior region where fractures are most commonly found. An error in the seed localization may easily lead to a sub-optimal outcome in such approaches. Zhang et al.~\cite{zhang_2016} developed a regression forest-based landmark detector to localize CMF landmarks on the CBCT scans. To address the spatial coherence of landmarks, image segmentation was used as a helper. The authors obtained a mean digitization error less than 2 $mm$ for $15$ CMF landmarks. The following year, to reduce the mean digitization error further, Zhang et al.~\cite{zhang_2017} proposed a deep learning based joint CMF bone segmentation and landmarking strategy. A context guided multi-task fully convolutional neural (FCN) network was employed along with 3D displacement maps to perceive the spatial locations of the landmarks. A segmentation accuracy of $93.27\pm 0.97 \%$ and a mean digitization error of less than $1.5$ mm for identifying $15$ CMF landmarks was achieved. Further, a joint segmentation and landmark digitization framework was proposed, where two stages of FCN were cascaded to perform bone segmentation and landmark localization \cite{zhang2020context}. The major disadvantage of this (one of the state-of-the-arts) method was the memory constraint introduced by the redundant information in the $3D$ displacement maps such that only a limited number of the landmarks can be learned using this approach. Since the proposed strategy is based on joint segmentation and landmarking, it naturally shares other disadvantages of the segmentation based methods: frequent failures for very challenging cases. The landmark localization problem was solved using an object detection method, where region proposals were used to identify landmark locations and a coarse to fine method was used to achieve landmark localization \cite{chen2021fast}. It must be noted, that the method does not use the relationships between the anatomical landmarks in the CMF bones.  

Recently, we integrated the manifold information (geodesic) in a deep learning architecture to improve robustness of the segmentation based strategies for landmarking~\cite{ness}, and obtained promising results, significantly better than the state-of-the-art methods. We also noticed that there is still a room to improve landmarking process especially when pathology or bone deformation is severe. To fill this research gap, in this study, we take a radically different approach by learning landmark relationships without segmenting bones. We hypothesize that the inherent relation of the landmarks in the CMF region can be learned by a relational reasoning algorithm based on deep learning. Although our proposed algorithm stems from this unique need of anatomical landmarking, the core idea of this work is inspired from the recent studies in artificial intelligence (AI), particularly in robotics and physical interactions of human/robots with their environments, as described in the following with further details.

\subsubsection{Relational Reasoning}
The ability to learn relationship and infer reasons between entities and their properties is a central component of the AI field, however it has been proven to be very difficult to learn through neural networks until recently~\cite{raposo2017discovering, yan2021multi}. In 2009, Scarselli et al.~\cite{4700287} introduced a graph neural network (GNN) by extending the neural network models to process graph data which encoded relationship information of the objects under investigation. Li et al.~\cite{li2015gated} proposed a machine learning model based on gated recurrent units (\textit{GRUs}) to learn the distributed vector representations from heap graphs. Despite this increase in use and promising nature of the GNN architectures~\cite{GNN2}, there is a limited understanding for their representational properties, which is often a necessity in medical AI applications for their adoption in clinics.

\begin{figure}
    \centering
    \includegraphics[width=0.7\columnwidth]{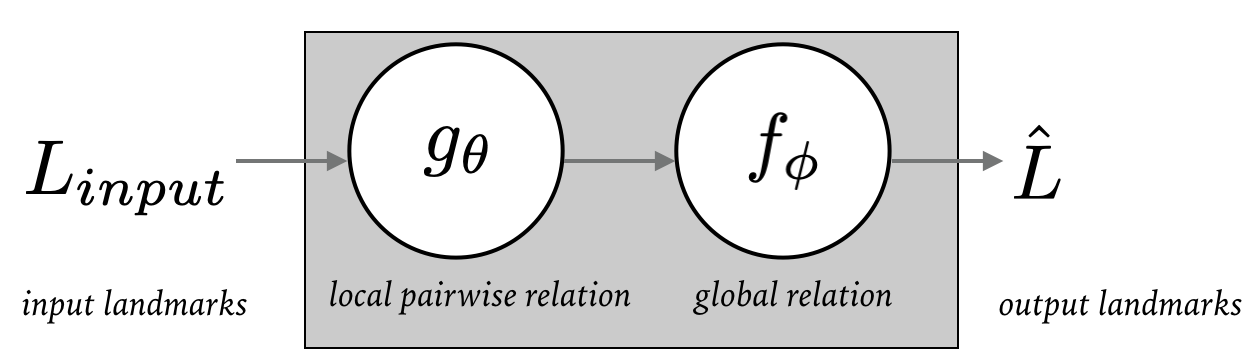}
    \caption{Overview of the proposed RRN architecture: for a few given input landmarks, RRN utilizes both pairwise and combination of all pairwise relations to predict the remaining landmarks.}
    \label{fig:arch_overview}
\end{figure}

Recently, DeepMind team(s) published four important studies on \textit{relational reasoning} and explored how objects in complex systems can interact with each other~\cite{relationalLearning_2,  raposo2017discovering, relationalLearning_1, relationalReasoning_3}. Battaglia et al.~\cite{relationalLearning_2} introduced interaction networks to \textit{reason} about the objects and the relations in the complex environments. The authors proposed a simple, yet accurate system to reason about $n$-body problems, rigid-body collision, and non-rigid dynamics. The proposed system can predict the dynamics in the next step with an order of magnitude lower error and higher accuracy. Raposa and Santoro et al.~\cite{raposo2017discovering} introduced a \textit{Relational Network (RN)} to learn the object relations from a scene description, hypothesising that a typical scene contains salient objects which are typically related to each other by their underlying causes and semantics. Following this study, Santoro and Raposa et al.~\cite{relationalLearning_1} presented another relational reasoning architecture for tasks such as visual question-answering, text-based question-answering, and dynamic physical systems where the proposed model obtained most answers correctly. Lastly, Battaglia et al.~\cite{relationalReasoning_3} studied the relational inductive biases to learn the relations of the entities and presented the graph networks. These four studies show promising approaches to understanding the challenge of relational reasoning. To the best of our knowledge, such advanced reasoning algorithms have neither been developed for nor applied to the medical imaging applications yet. It must be noted that medical AI applications require fundamentally different reasoning paradigms than conventional computer vision and robotics fields have \cite{zhou2018temporal} (e.g., salient objects definitions). To address this gap, in this study we focus on the anatomy-anatomy and anatomy-pathology relationships in an implicit manner. 

\subsection{Summary of our contributions}
\begin{itemize}
    \item  The proposed method is the first in the literature to successfully apply spatial reasoning of the anatomical landmarks for accurate and robust landmarking using deep learning. 
   \item Many anatomical landmarking  methods, including our previous works~\cite{ness, automated_landmarking_challanges_1, seg_landmarking}, use bone segmentation as a guidance for finding the location of the landmarks on the surface of a bone. The major limitation imposed by such an approach stems from the fact that it is not always possible to have an accurate segmentation. Our proposed RRN system addresses this problem by enabling accurate prediction of anatomical landmarks without employing explicit object segmentation. 
    \item Since efficiency is a significant barrier for many medical AI applications, we explore new deep learning architecture designs for a better efficacy in the system performance. For this purpose, we utilize variational dropout~\cite{Kingma:2015:VDL:2969442.2969527} and targeted dropout~\cite{gomez2019learning} in our implementation for faster and more robust convergence of the landmarking procedure ($\sim5$ times faster than baselines). 
     \item Our data set  includes  highly  variable  bone  deformities along  with  other  challenges  of  the  CBCT  scans with a larger number of scans (as compared to baselines).  Hence, the proposed algorithm is considered robust and identifies anatomical landmarks accurately under varying conditions (Table~\ref{table:experimental_results}). In our experiments, we find landmarks pertaining to mandible, maxilla, and nasal bones (Figure~\ref{fig:madible_maxilla}).
\end{itemize}

The rest of this paper is organized as follows: we introduce our novel methodology and its details in Section~\ref{section_2}. In Section~\ref{section_3}, we present experiments and results and then we conclude the paper with discussing strengths and limitations of our study in Section~\ref{section_4}. 

\begin{figure*}[h]
    \includegraphics[width=0.19\linewidth]{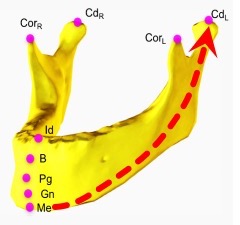}
    \includegraphics[width=0.19\linewidth]{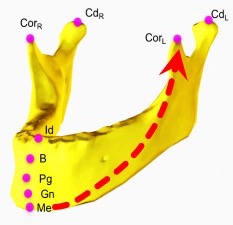}
    \includegraphics[width=0.19\linewidth]{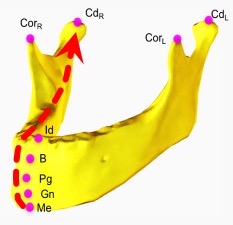}
    \includegraphics[width=0.19\linewidth]{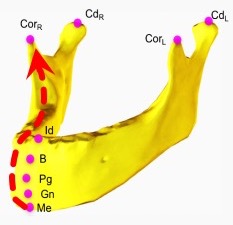}
    \includegraphics[width=0.19\linewidth]{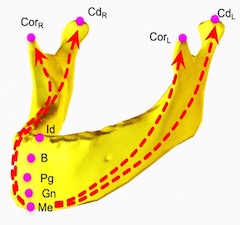}
    \caption{For the input domain $L_{input}=\{Me, Cd_L, Cor_L, Cd_R, Cor_R\}$: (a)-(d) pairwise relations of landmark \textit{Menton} a) \textit{Menton}-\textit{Condylar Left}, b)  \textit{Menton}-\textit{Coronoid Left}, c) \textit{Menton}-\textit{Condylar Right}, d) \textit{Menton}-\textit{Coronoid Right}, e) combined relations of \textit{Menton}.}
    \label{fig:pairwise}
\end{figure*}

\section{Methods}
\label{section_2}
\subsection{Overview and preliminaries}
The most frequently deformed or injured CMF bone is the lower jaw bone, \textit{mandible}, which is the only mobile CMF bone~\cite{ARMOND2017716}. In our previous study~\cite{ness}, we developed a framework to segment mandible from CBCT scans and identify the mandibular landmarks in a fully-automated way. Herein, we focus on anatomical landmarking without the need for explicit segmentation, and extend the learned landmarks into other bones (maxilla and nasal). Overall, we seek answers for the following important questions:

\begin{itemize}
 \item \textbf{Q1:} Can we automatically identify all anatomical landmarks of a bone if some of the landmarks are missing?  If so, what is the least effort for performing this procedure? How many landmarks are necessary and which landmarks are more informative to perform this whole procedure?
\item \textbf{Q2:} Can we identify anatomical landmarks of nasal and maxilla bones if we only know locations of a few landmarks in the mandible and the rest is missing? Do relations of landmarks hold true even when they belong to different anatomical  structures (manifold)?
\end{itemize}

In this study, we explore inherent relations among anatomical landmarks at the local and global levels in order to explore availability of structured data samples helping anatomical landmark localization. Inferred from the morphological integration of the CMF bones, we claim that landmarks of the same bone should carry common properties of the bone so that one landmark should give clues about the positions of the other landmarks with respect to a common reference. This reference is often chosen as segmentation of the bone to enhance information flow, but in our study we leverage this reference point from the whole segmented bone into a reference landmark point. Throughout the text, we use the following definitions:

\textbf{\textit{Definition 1:}} A \textit{landmark} is an anatomically distinct point, helping clinicians to make reliable measurements related to a condition, diagnosis, modeling a surgical procedure, or creating a treatment plan. 

\textit{\textbf{Definition 2:} }A \textit{relation} is defined as a geometric property between landmarks. Relations might include the following geometric features: size, distance, shape, and other implicit structural information. In this study, we focus on pairwise relations between landmarks as a starting point.

\textit{\textbf{Definition 3:}} A \textit{reason} is defined as an inference about relationships of the landmarks. For instance, compared to closely localized landmarks (if given as input), a few of sparsely localized landmarks can help predicting landmarks better. The reason is that sparsely localized input landmark configuration captures the anatomy of a region of interest and infers better global relationships of the landmarks. 

Once relationship among landmarks is learned effectively, we can use this relationship to identify the missing landmarks on the same or different CMF bones without the need for a precise segmentation. Towards this goal, we propose to learn a relationship of the anatomical landmarks in two stages (illustrated in Figure~\ref{fig:arch_overview}) based on \textit{Relational Units (RUs)}. The first stage which is shown as the function $g$, which learns the pairwise (local) relations. The second stage which is shown as a function $f$, which combines pairwise relations ($g$) of the first stage into a global relation based on \textit{RUs}.

Figure~\ref{fig:pairwise} shows example pairwise relations for different pairs of mandible landmarks. There are five sparsely localized landmarks. The basis/reference is chosen  as Menton, in this example, hence, four pairwise relations are illustrated from a to d. Figure~\ref{fig:pairwise}e illustrates combined relations (a-d) of the landmark Menton (reference) with respect to other four landmarks on the mandible.

\subsection{Relational Reasoning Architecture}
Anatomical landmarking has been an active research topic for several years in the medical imaging field. However, how to build a reliable/universal relationship between landmarks for a given clinical problem is an open problem. While anatomical similarities at the local and global levels could serve towards viable solutions, thus far, features that can represent anatomical landmarks from the medical images have not achieved the desired efficacy and interpretation~\cite{zhang_2017, ulas_bagci, gc_asm, RUEDA2010324}.

\begin{figure}[h]
\centering
    \includegraphics[width=0.8\linewidth]{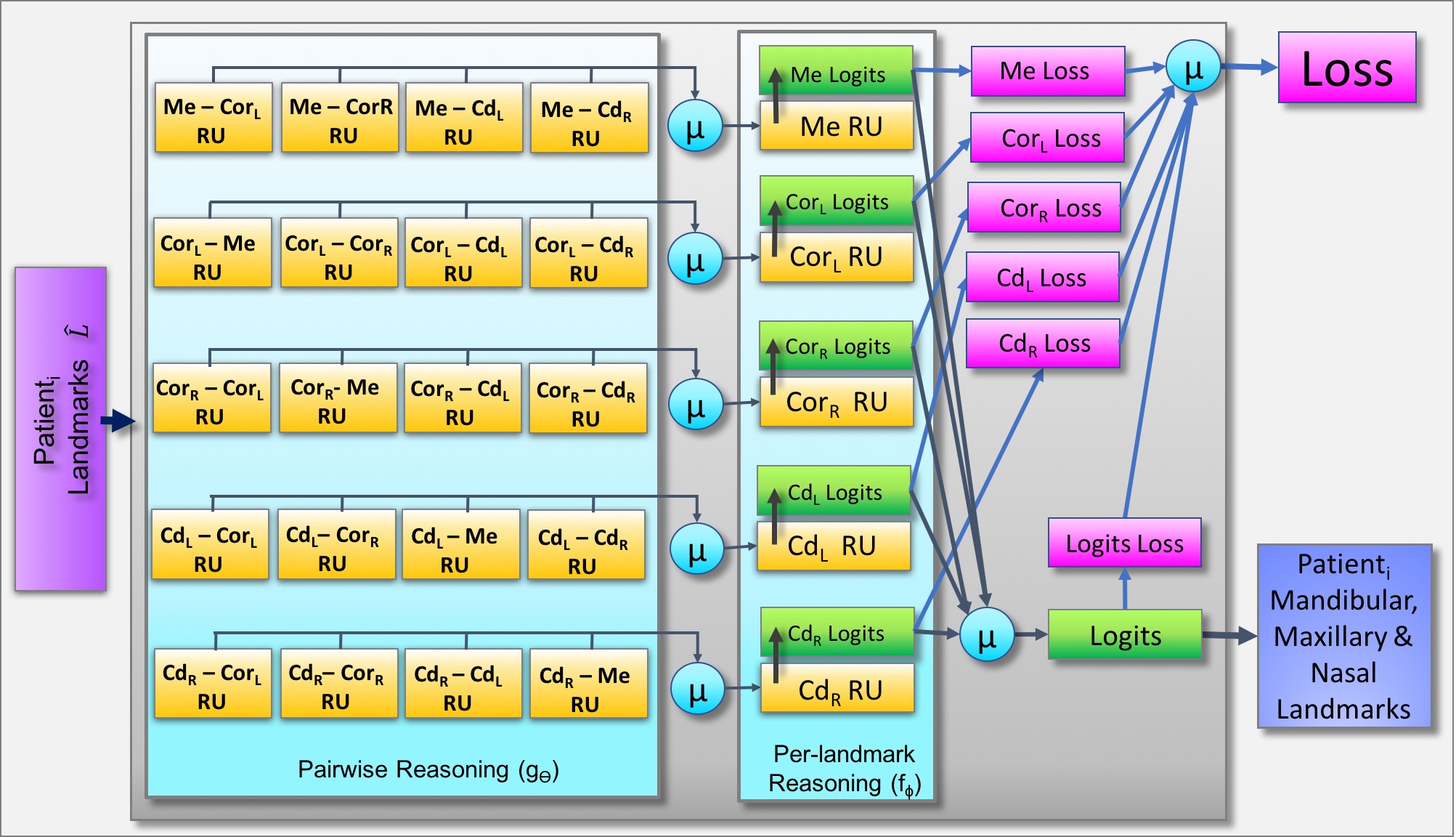}
    \includegraphics[width=0.4\linewidth]{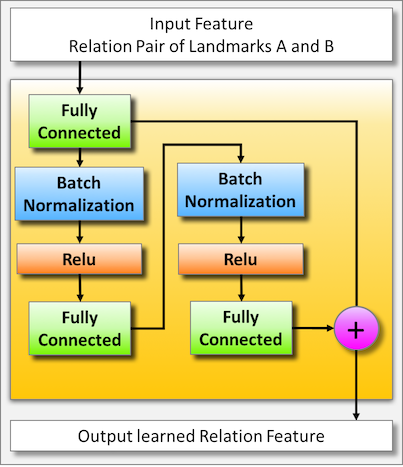}
    \includegraphics[width=0.23\linewidth]{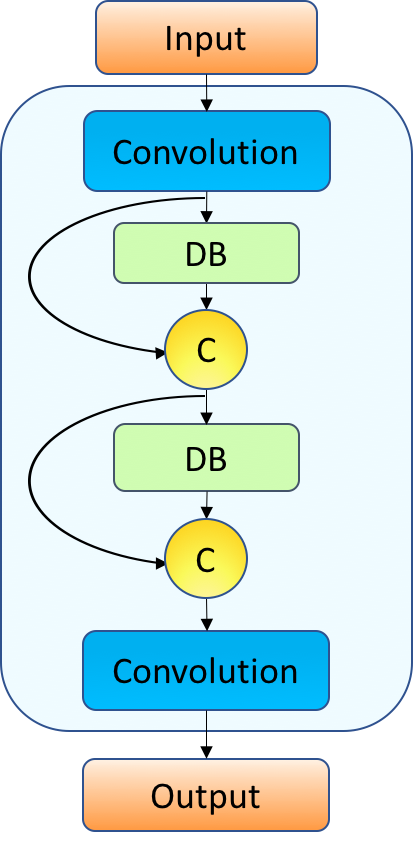}
    \includegraphics[width=0.3\linewidth]{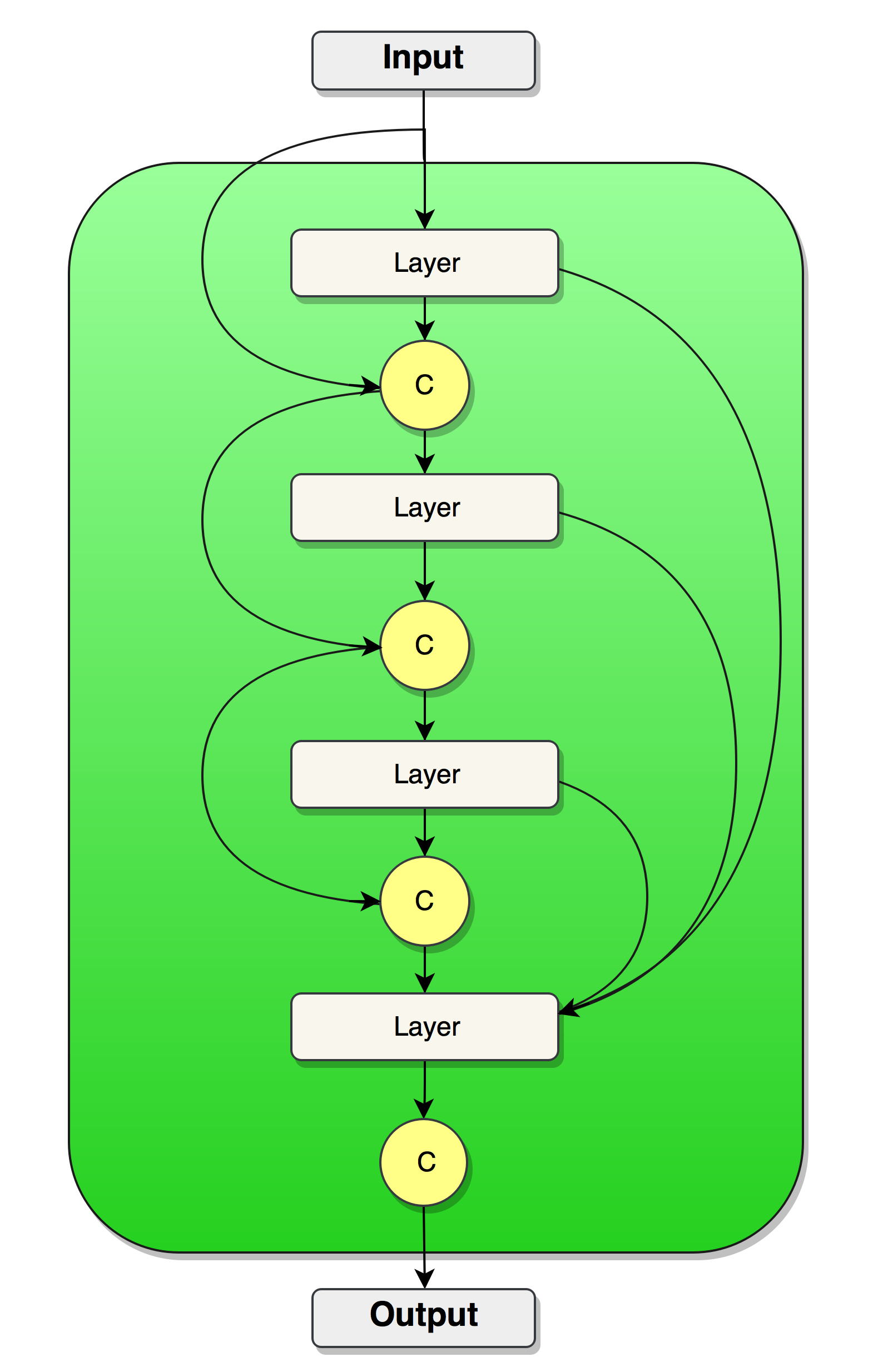}
     \caption{Top) RRN Architecture for $5$-input landmarks $RRN(L_{input})$:  $L_{input}$=$\{Me, Cor_L, Cor_R, Cd_L, Cd_R\}$, $\hat{L}=\{Gn, Pg,B, Id, Ans, A, Pr,Pns, Na\}$ and $\mu$ is the average operator. Bottom Left) Content of the pairwise reasoning block, Relation Unit (\textit{RU}) of the RRN. Bottom Middle) RU composed of 2 \textit{DBs}, convolution and concatenation (C) units. Bottom Right) Dense Block (\textit{DB}) architecture composed of $4$ layers and concatenation layers.
     }
    \label{fig:RU_RNN}
\end{figure}

We propose a new framework called relational reasoning network (\textit{RRN}) to learn local and global relations of anatomical landmarks ($o_i$) through its units called \textit{RU} (relationship unit). The proposed \textit{RRN} architecture and its \textit{RU} sub-architectures are summarized in Figure~\ref{fig:RU_RNN}. The relation of two landmarks are encoded via major spatial properties of the landmarks. We explore two architectures as \textit{RU}: first one is a simple multi-layer perceptron (\textit{MLP}) (Figure~\ref{fig:RU_RNN}-Bottom Left) (similar to~\cite{raposo2017discovering}), the other one is more advanced architecture composed of Dense-Blocks (\textit{DBs}) (Figure~\ref{fig:RU_RNN} Bottom Middle). Both architectures are relatively simple compared to very dense complex deep-learning architectures. Our objective is to locate all anatomical landmarks by inputting a few landmarks to \textit{RRN}, which provides reasoning inferred from the learned relationships of landmarks and locate all other landmarks automatically. 

 In the pairwise learning/reasoning stage (stage 1), 5-landmarks based system is assumed as an example network (other configurations are possible too, see experiments and results section). Sparsely-spaced landmarks (Figure~\ref{fig:pairwise}e) and their pairwise relationships are learned in this stage ($g_{\theta}$). These pairwise relationship(s) are later combined in a separate \textit{DB} setting in ($f_{\phi}$). It should be noted that this combination is employed through a joint loss function and an \textit{RU} to infer an average relation information. In other words, for each individual landmark, the combined relationship vector is assigned a secondary learning function through a single \textit{RU}. 

The \textit{RU} is the core component of the \textit{RRN} architecture. Each \textit{RU} is designed in an end-to-end fashion; hence, they are differentiable. For $n$ landmarks in the input domain, the proposed \textit{RRN} architecture learns $n \times (n-1)$ pairwise and $n$ combined relations (global) with a total of $n^2$ \textit{RUs}. Therefore, depending on the number of input domain landmarks, \textit{RRN} can be either shallow or dense. Let $L_{input}$ and $\hat{L}$ indicate vectors of input and output anatomical landmarks, respectively. Then, two stages of the \textit{RRN} of the input domain landmarks $L_{input}$ can be defined as:

\begin{equation}
\begin{aligned}
   {G_{\theta}}_i=\frac{1}{(n-1)}\sum_{j=1,j \neq i}^{n}(g_\theta(o_i,o_j)), \\
   RRN(L_{input};\theta,\phi) =  \frac{1}{n}\sum_{i=1}^{n}{f_{\phi}}_i({G_{\theta}}_i),
   \label{eq:RNN}
\end{aligned}
\end{equation}

\noindent where ${G_{\theta}}_i$ is the mean pairwise relation vector  of the landmark $o_i$ to every other landmark $o_{j(j \neq i)}  \in L_{input}$. The functions $f_{\phi}$ and $g_{\theta}$ are the functions with the free parameters $\phi$ and $\theta$, and $f_{\phi}$ indicates a global relation (in other words, combined pairwise relations) of landmarks. 

\subsection{Pairwise Relation ($g_{\theta}$)}
For a given a few input landmarks ($L_{input}$),  our objective is to predict the $3D$ spatial locations of the target domain landmarks ($\in \hat{L}$) by using the $3D$ spatial locations of the input domain landmarks ($\in L_{input}$). With respect to relative locations of the input domain landmarks, we reason about the locations of the target domain landmarks. The \textit{RU} function $g_{\theta}(o_i,o_j)$ represents the relation of two input domain landmarks $o_i$ and $o_j$ where $i \neq j$ (Figure~\ref{fig:pairwise}a-d). The output of $g_{\theta}(o_i,o_j)$ describes relative spatial context of two landmarks, defined for each pair of input domain landmarks (pairwise relation at Figure~\ref{fig:RU_RNN}). According to each input domain landmark $o_i$, the structure of the manifold is captured through mean of all pairwise relations  (represented as $G_{\theta_{i}}$ at Equation~\ref{eq:RNN}). 


\subsection{Global Relation ($f_{\phi}$)}
The mean pairwise relation  ${G_{\theta}}_i$ is calculated with respect to each input domain landmark $o_i$, and it is given as input to the second stage where global (combined) relation ${f_{\phi}}_i$ is learned. ${f_{\phi}}_i$ is a \textit{RU} function and the output of ${f_{\phi}}_i$ is the predicted $3D$ coordinates of the target domain landmarks ($\in \hat{L}$). In other words, each input domain landmark $o_i$ learns and predicts the target domain landmarks by the \textit{RU} function ${f_{\phi}}_i$. The terminal prediction of the target domain landmarks is the average of individual predictions of each input domain landmark, represented by $RRN(L_{input};\theta,\phi)$ at Equation~\ref{eq:RNN}. There are totally $n^2$ \textit{RUs} in the architecture. The number of trainable parameters used for each experimental configuration are directly proportional with $n^2$ (Table~\ref{table:setip_input_output}).  Since all pairwise relations are leveraged under $G_{\theta_{i}}$ and ${f_{\phi}}$ with averaging operation, we can conclude that RRN is invariant to the order of input landmarks (i.e., permutation-invariant). 


\subsection{Loss Function}
The natural choice for the loss function is the mean squared error (\textit{MSE}) because it is a differentiable distance metric measuring how well landmarks are localized/matched, and it allows output of the proposed network to be real-valued functions of the input landmarks. For $n$ input landmarks and $m$ target landmarks, \textit{MSE} simply penalizes large distances between the landmarks as follows:

\begin{equation}
   Loss\left(W_{\Theta},(\theta,\phi)\right) = \frac{1}{n*m}\sum_{i=1}^{n}\left(\sum_{k=1}^{m}||{(f_{\phi}({G_{\theta}}_i))}_k-o_k||^2\right)  
   \label{eq:Loss}
\end{equation}
\noindent where ${o_k}$ are target domain landmarks ($o_k\in \hat{L})$.

\subsection{Variational Dropout}
Dropout is an important regularizer employed to prevent overfitting at a cost of $2-3$ times (on average) increase in training time~\cite{JMLR:v15:srivastava14a}. For efficiency reasons, speeding up dropout is critical and it can be achieved by a variational Bayesian inference on the model parameters~\cite{Kingma:2015:VDL:2969442.2969527}. Given a training input dataset $X=\{x_1,x_2,..,x_N\}$ and the corresponding output dataset $Y=\{y_1,y_2,..,y_N\}$, the goal in \textit{RRN} is to learn the parameters $\omega$ such that $y=F_{\omega}(x)$. In the Bayesian approach, given the input and output datasets $X,Y$, we seek for the posterior distribution $p(\omega|X,Y)$, by which we can predict output $y^*$ for a new input point $x^*$ by solving the integral \cite{Gal:2016:TGA:3157096.3157211}:

\begin{equation}
    p(y^*|x^*,X,Y) = \int p(y^*|X^*,\omega)p(\omega|X,Y)d\omega.
\end{equation}

In practice, this computation involves intractable integrals \cite{Kingma:2015:VDL:2969442.2969527}. To obtain the posterior distributions, a Gaussian prior distribution $N(0, I)$ is placed over the network weights \cite{Gal:2016:TGA:3157096.3157211} which leads to a much faster convergence \cite{Kingma:2015:VDL:2969442.2969527}.

\subsection{Targeted Dropout}
Alternatively, we also propose to use targeted dropout for better convergence~\cite{gomez2019learning}. Given a neural network parameterized by $\Theta$, the goal is to find the optimal parameters $W_{\Theta}(.)$ such that the loss $Loss(W_{\Theta})$ is minimized. For efficiency and generalization reasons, $|W_{\Theta}| \leq k $, only $k$ weights of highest magnitude in the network are employed. In this regard, deterministic approach is to drop the lowest $|W_{\Theta}|-k$ weights. In targeted dropout, using a target rate $\gamma$ and a drop out rate $\alpha$, first a target set is generated with the lowest weights with the target rate $\gamma$. Next, weights are dropped out in an stochastic manner from the target set at a certain dropout rate $\alpha$. 

\subsection{Landmark Features}
Pairwise relations are learned through \textit{RU} functions. Each \textit{RU} accepts input features to be modelled as a pairwise relation. It is desirable to have such features characterizing landmarks and interactions with other landmarks. These input features can either be learned throughout a more complicated network design, or through feature engineering. 
In this study, for simplicity, we define a set of simple yet explainable geometric features. Since \textit{RUs} model relations between two landmarks ($o_A$ and $o_B$), we use  $3D$ coordinates of these landmarks (both in pixel and spherical space), their relative positions with respect to a well-defined landmark point (reference), and  approximate size of the mandible.  The mandible size is estimated as the distance between the maximum and the minimum coordinates of the input domain mandibular landmarks (Table~\ref{table:stage1_input_feature}). Finally, a $19$-dimensional feature vector is considered to be an input to local relationship function $g$. For a well-defined reference landmark, we used \textit{Menton} \textit{(Me)} as the origin of the Mandible region.

\begin {table}[t]
\centering
\caption{Input landmarks have the following feature(s) to be used only in stage I.  $19D$ feature vector includes only structural information.}
\label{table:stage1_input_feature}
\begin{tabular}{p{7cm}|p{5cm}} 

\multicolumn{2}{c}{Pairwise Feature ($o_{A}$, $o_{B}$)} \\ \hline
\makecell[l]{3D pixel-space position of the $o_A$} & \makecell[c]{$(A_{x}, A_{y}, A_{z}$)} \\ \hline
\makecell[l]{Spherical coordinate of the vector \\from landmark Menton ($o_1$) to $o_A$} & \makecell[c]{$(r_{me \rightarrow A}$, $\theta_{me \rightarrow A}$, $\phi_{me \rightarrow A}$)} \\ \hline
3D pixel-space position of the $o_B$ &\makecell[c]{ $(B_{x}, B_{y}, B_{z}$)} \\ \hline 
\makecell[l]{Spherical coordinate of the vector \\from landmark Menton to $l_B$} & \makecell[c]{($r_{me \rightarrow B}$, $\theta_{me \rightarrow B}$, $\phi_{me \rightarrow B}$)} \\ \hline
\makecell[l]{3D pixel-space position of the \\landmark Menton} & \makecell[c]{$(Me_{x}, Me_{y}, Me_{z}$)} \\ \hline
\makecell[l]{Spherical coordinate of the vector \\from $o_A$ to $o_B$} &  \makecell[c]{($r_{A \rightarrow B}$, $\theta_{A \rightarrow B}$, $\phi_{A \rightarrow B}$)} \\ \hline
\makecell[l]{Diagonal length of the bounding box \\capturing Mandible roughly, computed\\as the distance between the minimum\\and the maximum spatial  locations\\of the input domain mandibular\\landmarks ($L_1$) in the pixel space.} & \thead{$d_{1}$} \\ \hline
\end{tabular}
\end{table}

\section{Experiments and Results}
\label{section_3}

\subsection{Data Description}
Anonymized CBCT scans of 250 patients (142 female and 108 male, mean age = 23.6 years, standard deviation = 9.34 years) were included in our analysis through an IRB-approved protocol. The data set includes both pediatric and adult patients with craniofacial congenital birth defects, developmental growth anomalies, trauma to the CMF, and surgical interventions. CB MercuRay CBCT system (Hitachi Medical Corporation, Tokyo, Japan) was used to scan the data  at $10$ mA and $100$ Kvp. The radiation dosage for each scan was around $300$ mSv. To handle the computational cost, each patient's scan was re-sampled from $512 \times 512 \times 512$ to $256 \times 256 \times 512$. In-plane resolution of the scans were noted (in mm) either as $0.754 \times 0.754 \times 0.377$ or $0.584\times 0.584 \times 0.292$.  In addition, following image-based variations exist in the data set: aliasing artifacts due to braces, metal alloy surgical implants (screws and plates), dental fillings, and missing bones or teeth~\cite{ness}.

The data was annotated independently by three expert interpreters, one from the NIH team, and two from UCF team. Inter-observer agreement values were computed as approximately $3$ pixels. Experts used freely available $3D$ Slicer software for the annotations~\cite{ness}. 

\begin {table*}[t]
\centering
\caption{Five Experimental Landmark Configurations for experimental explorations. $L_{input}$: input landmarks and $\hat{L}$: output landmarks, and $\#$\textit{RUs} indicate the number of relational units. Landmarks are visualized using reference standard bones for illustrative purposes; in our implementation there is no explicit segmentation exist.}
\label{table:setip_input_output}
\resizebox{\linewidth}{!}{%
\begin{tabular}{c|c|c|c|c|c|c|c|c}
\toprule
\textbf{\thead{Experiment \#}} & \textbf{Configuration} & \multicolumn{3}{c|}{\thead{$\mathbf{L_{input}}$}} & \multicolumn{3}{c|}{\thead{$\mathbf{\hat{L}}$}} & $\#$\textit{RUs} \\
\midrule
\textbf{1} & \textbf{$\mathbf{5}$-landmarks} & \thead{$Me$, $Cd_L$,\\$Cd_R$, $Cor_L$,\\$Cor_R$} & \multicolumn{2}{c|}{\raisebox{-.5\height}{\includegraphics[height=20mm]{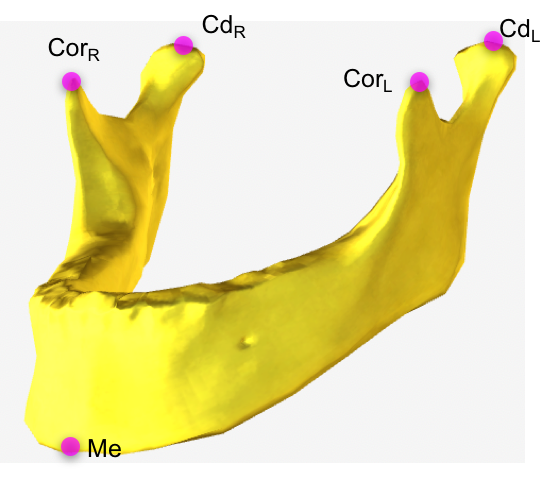}}} & \thead{$Gn$,$Pg$,$B$, \\ $Id$,$Ans$,$A$, \\ $Pr$,$Pns$, $Na$}  &
\raisebox{-.5\height}{\includegraphics[height=20mm]{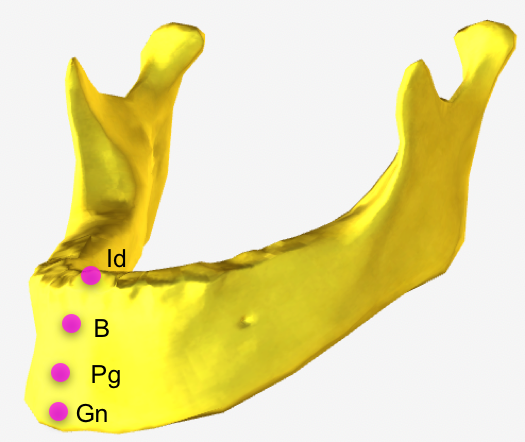}} &
\raisebox{-.5\height}{\includegraphics[height=20mm]{images/maxilla_landmarks_2.png}} & 25 \\ \hline 
\textbf{2} & \textbf{\thead{$\mathbf{3}$-Landmarks \\ Regular}} & \thead{$Me$, $Cd_L$,\\$Cd_R$} & \multicolumn{2}{c|}{\raisebox{-.5\height}{\includegraphics[height=20mm]{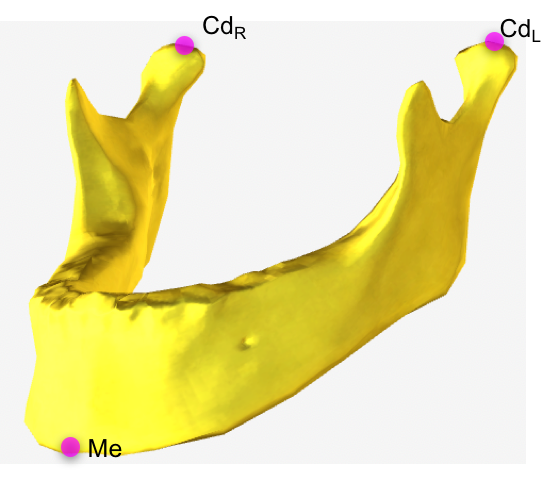}}} &  \thead{$Gn$,$Pg$,$B$, $Id$, \\ $Cor_L$, $Cor_R$, \\$Ans$,$A$,$Pr$, \\ $Pns$, $Na$}   & \raisebox{-.5\height}{\includegraphics[height=20mm]{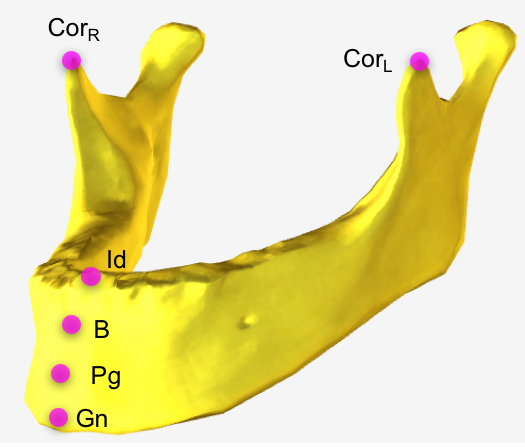}} & \raisebox{-.5\height}{\includegraphics[height=20mm]{images/maxilla_landmarks_2.png}} & 9 \\ \hline
\textbf{3} & \textbf{\thead{$\mathbf{3}$-Landmarks \\ Cross}} & \thead{$Me$, $Cd_R$,\\$Cor_L$} & \multicolumn{2}{c|}{\raisebox{-.5\height}{\includegraphics[height=20mm]{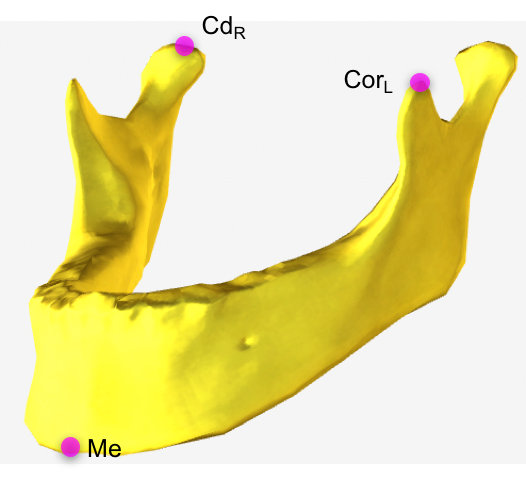}}} & \thead{$Gn$,$Pg$,$B$, \\ $Id$, $Cd_L$, $Cor_R$,\\ $Ans$,$A$,$Pr$, \\ $Pns$, $Na$}  & \raisebox{-.5\height}{\includegraphics[height=20mm]{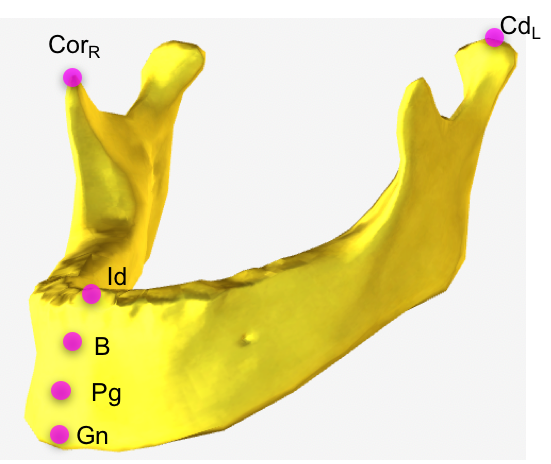}} & \raisebox{-.5\height}{\includegraphics[height=20mm]{images/maxilla_landmarks_2.png}} & 9 \\ \hline 
\textbf{4} & \textbf{$\mathbf{6}$-landmarks} & \thead{$Me$, $Cd_L$,\\$Cd_R$, $Cor_L$,\\$Cor_R$, $Na$} &
\raisebox{-.5\height}{\includegraphics[height=20mm]{images/baselineLandmarks.png}} &  \raisebox{-.5\height}{\includegraphics[height=20mm]{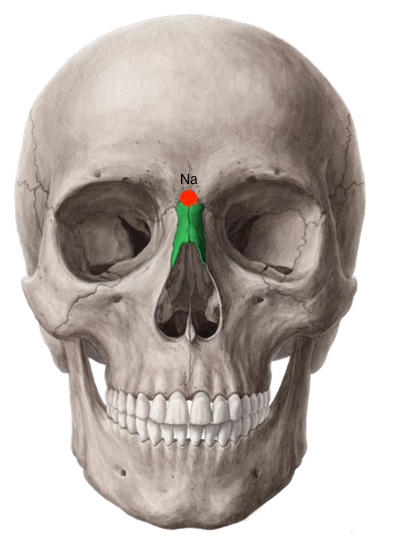}} & \thead{$Gn$,$Pg$,$B$, \\ $Id$,$Ans$,$A$, \\ $Pr$,$Pns$} & \raisebox{-.5\height}{\includegraphics[height=20mm]{images/baseline_out.png}} & \raisebox{-.5\height}{\includegraphics[height=20mm]{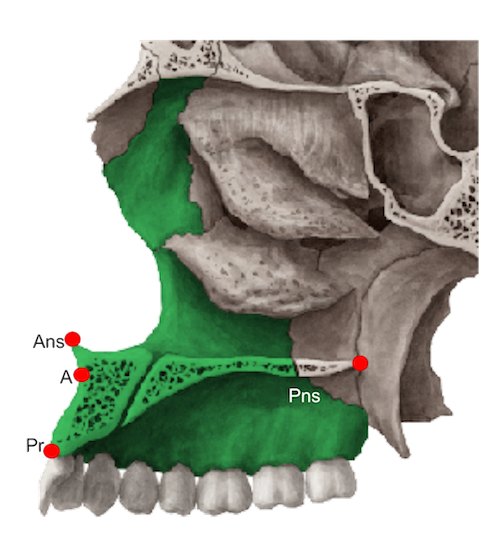}} & $36$ \\ \hline 
\textbf{5} & \textbf{$\mathbf{9}$-landmarks} & \thead{$Me$, $Cd_L$,\\$Cd_R$, $Cor_L$,\\$Cor_R$, $Gn$,\\$Pg$, $B$, $Id$} & \multicolumn{2}{c|}{\raisebox{-.5\height}{\includegraphics[height=20mm]{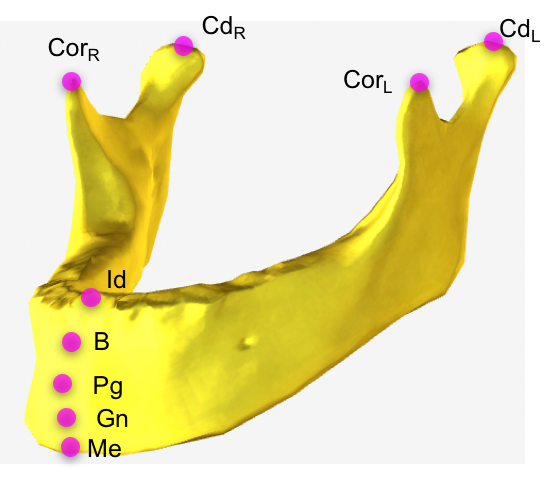}}} & \thead{$Ans$,$A$,$Pr$, \\ $Pns$, $Na$} & \multicolumn{2}{c|}{ \raisebox{-.5\height}{\includegraphics[height=20mm]{images/maxilla_landmarks_2.png}}} & 81 \\ \hline 
\end{tabular}}
\end{table*}

\subsection {Data Augmentation}
Our data set includes fully-annotated mandibular, maxillary, and nasal bones' landmarks. Due to insufficiency of $250$ samples for a deep-learning algorithm to run, we applied data-augmentation approach. In our study, the common usage of random scaling or rotations for data-augmentation were not found to be useful for new landmark data generation because such transformations would not generate new relations different from the original ones. Instead, we used random interpolation similar to active shape model's landmarks~\cite{doi:10.1118/1.3602070}. Briefly, we interpolated $2$ (or $3$) randomly selected scans with randomly computed weight. We merged the relation information at different scans to a new relation. We also added random noise to each landmark with a maximum in the range of $\pm 5$ pixels, defined empirically based on the resolution of the images as well as the observed high-deformity of the bones. We generated a dataset with approximately $100K$ landmarks, which is empirically evaluated as a sufficiently large dataset.

\subsection{Evaluation Methods}
We used root-mean squared error (\textit{RMSE}) in the anatomical space (in mm) to evaluate the goodness of the landmarking. Lower \textit{RMSE} indicates successful landmarking process. For statistical significance comparisons of different methods and their variants, we used a P-value of $0.05$ as a cut-off threshold to define significance and applied $t$-tests where applicable. 

\subsection{Input Landmark Configurations}

In our experiments, there were three groups of landmarks (See Figure~\ref{fig:madible_maxilla}) defined based on the bones they reside: Mandibular $L_1=\{o_1,...,o_9\}$, Maxillary $L_2=\{o_{10},..., o_{13}\}$, and Nasal $L_3=\{o_{14}\}$, where subscripts in $o$ denote the specific landmark in that bone: 
\begin{itemize}
    \item $L_1=\{Me, Gn, Pg, B, Id, Cor_L, Cor_R, Cd_L, Cd_R\}$, 
    \item $L_2=\{Ans, A, Pr, Pns,\}$,
    \item $L_3=\{Na\}$.
\end{itemize}

In each experiment, as detailed in Table~\ref{table:setip_input_output}, we designed a specific input set $L_{input}$ where $L_{input} \subseteq L_1 \cup L_2$, $|L_{input}|=n$ and $1<n<=(|L_1|+|L_2|)$. 
The target domain landmarks for each experiment were $\hat{L}=(L_1 \space \cup \space L_2 \space \cup \space L_3 \space) \setminus \space L_{input}$ and $|\hat{L}|=m$ such that $n+m=14$. With carefully designed input domain configurations $L_{input}$, and pairwise relationships of the landmarks in the input set, we seek the answers to the following questions previously defined as Q1 and Q2 in Section~\ref{section_2}:

\begin{itemize}
    \item What configuration of the input landmarks can capture the manifold of bones so that other landmarks can be localized successfully?
    \item What is the minimum number and configuration of the input landmarks for successful identification of other landmarks?
\end{itemize}

Overall, we designed $5$ different input landmark configurations called $3$-landmarks regular, $3$-landmarks cross, $5$-landmarks, $6$-landmarks, and $9$-landmarks (Table~\ref{table:setip_input_output}).  Each configuration is explained in Section~\ref{subsection_exp}. 

\subsection{Training}

The \textit{MLP} \textit{RU} was composed of $3$ fully-connected layers, $2$ batch normalizations and $2$ ReLUs (Figure~\ref{fig:RU_RNN}). The \textit{DB} \textit{RU} architecture contained $1$ \textit{DB}, which was composed of $4$ layers with a growth-rate of $4$. We used a batch size of $64$ for all experiments. For the $5$-landmarks configuration, there were $6,596,745$ and $11,068,655$ trainable parameters for the \textit{MLP} and the \textit{DB} architectures, respectively. We trained the network for $100$ epochs on $1$ Nvidia Titan-XP GPU with $12GB$ memory using the \textit{MLP} architecture with the regular dropout compared to $20$ epochs with the variational and targeted dropout implementations. For the \textit{DB} architecture, it converged in around $20$ epochs independent of the dropout implementation employed. 

\subsection{Experiments and Results}
\label{subsection_exp}

\begin{figure*}[t]
    \centering
    \begin{subfigure}[$\space\space$Patient-$1$]{
    \label{fig:p_1}
    \includegraphics[height=8.5cm]{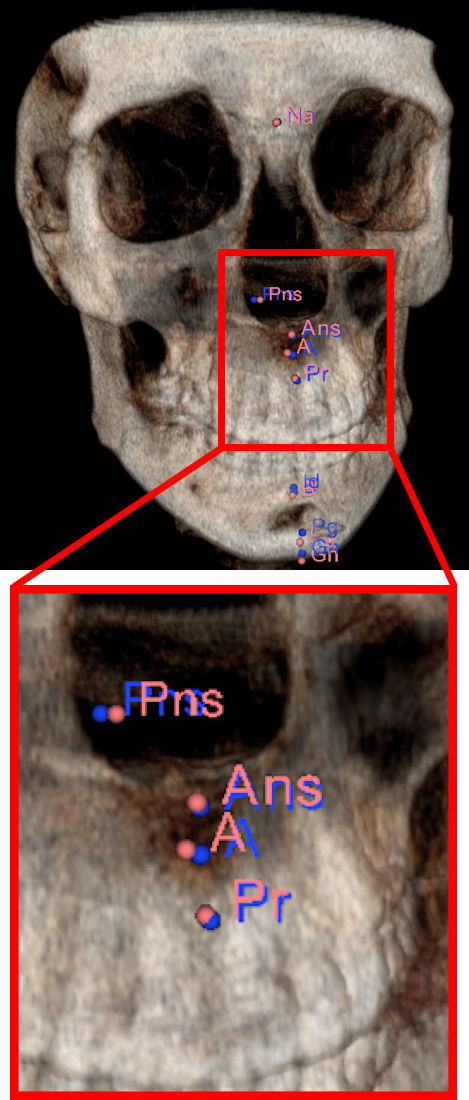}}
    \end{subfigure}
    \begin{subfigure}[$\space\space$Patient-$2$]{
    \label{fig:p_2}
    \includegraphics[height=8.5cm]{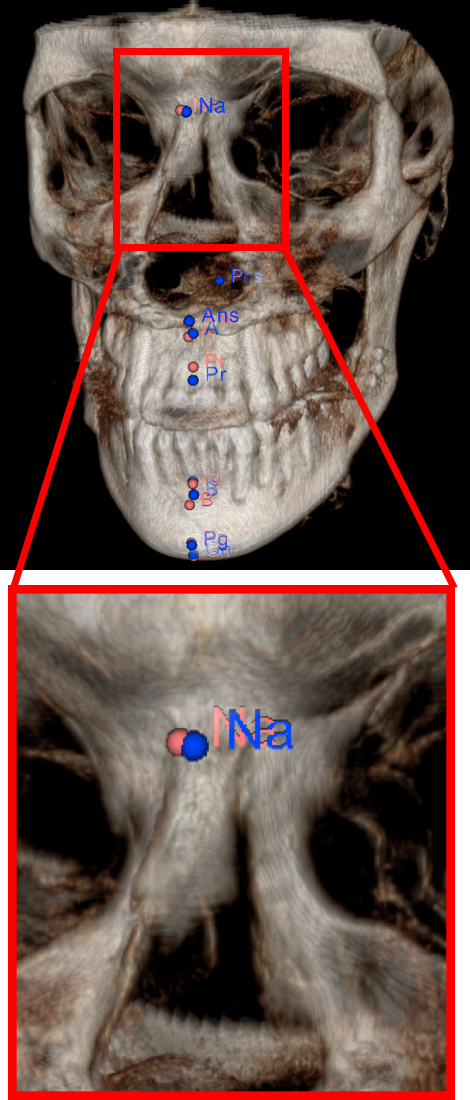}}
    \end{subfigure}
    \begin{subfigure}[$\space\space$Patient-$3$]{
    \includegraphics[height=8.5cm]{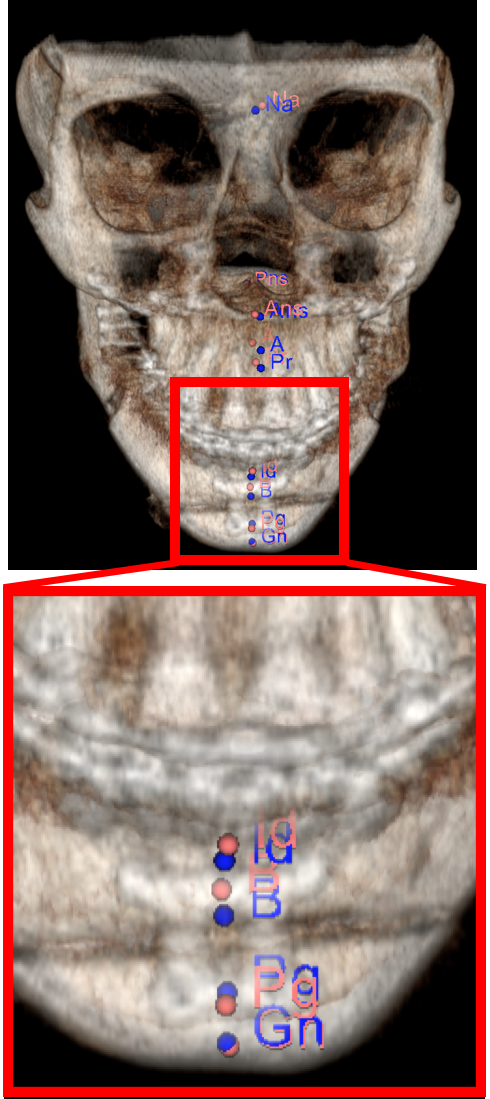}}
    \end{subfigure}
    \caption{Landmark annotations using the $5$-landmarks configuration: Ground truth in blue and computed landmarks in pink. a) Genioplasty/chin advancement (male 43 yo), b) Malocclusion (mandibular hyperplasia, maxillary hypoplasia) surgery (male 19 yo),  c) Malocclusion (mandibular hyperplasia,  maxillary hypoplasia) surgery (female 14 yo). Note that landmarks are shown on the volume-rendered CBCT scans, there is no segmentation conducted.}
    \label{fig:results}
\end{figure*}

We ran a set of experiments to evaluate the performance of the proposed system using a $4$-fold cross-validation. We summarized the experimental configurations in Table~\ref{table:setip_input_output}, error rates in Table~\ref{table:experimental_results}, and corresponding renderings in Figure~\ref{fig:results}. The method achieving the minimum error for a corresponding landmark is colored the same as the corresponding landmark at Table~\ref{table:experimental_results}. As shown by results, the minimum number of the input landmarks required for successful identification of other landmarks is determined as 3.

\renewcommand{\arraystretch}{1.05}

\begin {table*}[h]
\centering
\caption{Landmark Localization Errors (mm). The symbol '-' means not applicable (N/A).}
\label{table:experimental_results}
\resizebox{\linewidth}{!}{
\begin{tabular}{c|c|c|c|c|c|c|c}
\toprule
\multirow{2}{*}{\textbf{\thead{Method}}} & \multicolumn{7}{c}{Mandibular Landmarks} \\ \cline{2-8}
& $\mathbf{Cor_R}$ & $\mathbf{Cor_L}$ & $\mathbf{Cd_L}$ & $\mathbf{Gn}$ & $\mathbf{Pg}$ & $\mathbf{B}$ & $\mathbf{Id}$ \\
\midrule
\cellcolor{red!25}\textbf{$3$-Landmarks Regular (Dense)} & $3.32 \pm 0.30$ & $3.03 \pm 0.31$ & - & $0.01 \pm 0.03$ & $0.09 \pm 0.11$ & \cellcolor{red!25}$\mathbf{0.60 \pm 0.15}$ & $0.56 \pm 0.19$ \\
\textbf{\thead{$3$-Landmarks Cross (Dense)}} & $1.88 \pm 0.24$ & - & $1.70 \pm 0.23$ & $0.007 \pm 0.03$ & $0.10 \pm 0.11$ & $0.77 \pm 0.18$ & $0.58 \pm 0.20$ \\
\textbf{\thead{$5$-landmarks Var. Dropout (MLP)}} & - & - & - & $0.05 \pm 0.05$ & $0.22 \pm 0.13$ & $0.91 \pm 0.16$ & $0.95 \pm 0.19$ \\ 
\cellcolor{blue!25}\textbf{$5$-landmarks (Dense)} & - & - & - & \cellcolor{blue!25}$\mathbf{0.0002 \pm 0.03}$ & $0.13 \pm 0.11$ & $0.87 \pm 0.16$ & $0.78 \pm 0.19$ \\
\textbf{\thead{$5$-landmarks Var. Dropout (Dense)}} & - & - & - & $0.0008 \pm 0.02$ & $0.07 \pm 0.02$ & $0.76 \pm 0.10$ & $0.64 \pm 0.18$ \\
\cellcolor{yellow!25}\textbf{$5$-landmarks Targeted Dropout (Dense)} & - & - & - & $0.004 \pm 0.03$ & \cellcolor{yellow!25}$\mathbf{0.063 \pm 0.11}$ & $0.71 \pm 0.16$ & $0.64 \pm 0.20$ \\
\textbf{\thead{$6$-landmarks (Dense)}} & - & - & - & $1.52 \pm 0.30$ & $0.86 \pm 0.29$ & $1.07 \pm 0.25$ & $1.24  \pm 0.24$\\
\textbf{\thead{$6$-landmarks Var. Dropout (Dense)}} & - & - & - & $1.04 \pm 0.30$ & $1.18 \pm 0.30$ & $0.86 \pm 0.28$ & $1.06 \pm 0.24$\\
\textbf{\thead{$6$-landmarks Targeted Dropout(Dense)}} & - & - & - & $1.20 \pm 0.29$ & $0.92 \pm 0.28$ & $1.09 \pm 0.24$  & $1.21 \pm 0.25$ \\
\textbf{\thead{$9$-landmarks (Dense)}} & - & - & - & - & - & - & - \\
\cellcolor{green!25}\textbf{Torosdagli et al.~\cite{ness}} & \cellcolor{green!25}$\mathbf{0.03}$ & \cellcolor{green!25}$\mathbf{0.27}$ & \cellcolor{green!25}$\mathbf{1.01}$ & $0.41$ & $1.36$ & $0.68$ & \cellcolor{green!25}$\mathbf{0.35}$ \\
\textbf{\thead{Gupta et al.~\cite{automated_landmarking_challanges_1}}} & - & - & $3.20$ & $1.62$ & $1.53$ & $2.08$ & - \\
\midrule
\multirow{2}{*}{\textbf{\thead{Method}}} & \multicolumn{7}{c}{Maxillary-Nasal Bone Landmarks} \\ \cline{2-8}
& $\mathbf{Ans}$ & $\mathbf{A}$ & $\mathbf{Pr}$ & \multicolumn{2}{c|}{$\mathbf{Pns}$} & \multicolumn{2}{c}{$\mathbf{Na}$} \\ \hline
\textbf{\thead{$3$-Landmarks Regular (Dense)}} & $3.04 \pm 0.39$ & $3.04 \pm 0.40$ & $2.89 \pm 0.40$ & \multicolumn{2}{c|}{$2.04 \pm 0.29$} & \multicolumn{2}{c}{$3.15 \pm 0.34$} \\
\textbf{\thead{$3$-Landmarks Cross (Dense)}} & $3.18 \pm 0.39$ & $3.14 \pm 0.39$ & $3.17 \pm 0.38$  & \multicolumn{2}{c|}{$2.61 \pm 0.33$} & \multicolumn{2}{c}{$3.13 \pm 0.37$} \\
\textbf{\thead{$5$-landmarks Var. Dropout (MLP)}} & $3.80 \pm 0.44$ & $3.95 \pm 0.48$ & $3.06 \pm 0.01$ & \multicolumn{2}{c|}{$3.85 \pm 0.42$} & \multicolumn{2}{c}{$3.20 \pm 0.34$} \\ 
\textbf{\thead{$5$-landmarks (Dense)}} & $3.21 \pm 0.27$ & $3.16 \pm 0.41$ & $2.92 \pm 0.42$ & \multicolumn{2}{c|}{$2.37 \pm 0.35$} & \multicolumn{2}{c}{$2.91 \pm 0.40$} \\
\textbf{\thead{$5$-landmarks Var. Dropout (Dense)}} & $3.15 \pm 0.21$ & $3.07 \pm 0.38$ & $3.09 \pm 0.40$ & \multicolumn{2}{c|}{$2.35 \pm 0.32$} & \multicolumn{2}{c}{$3.14 \pm 0.36$} \\
\textbf{\thead{$5$-landmarks Targeted Dropout (Dense)}} &  $3.17 \pm 0.38$ & $3.09 \pm 0.39 $ & $2.85 \pm 0.39$ & \multicolumn{2}{c|}{$2.46 \pm  0.32$} & \multicolumn{2}{c}{$3.14 \pm 0.40$} \\
\cellcolor{red!25}\textbf{$6$-landmarks (Dense)} & $0.79 \pm 0.23$ & $1.65 \pm 0.29$ & \cellcolor{red!25}$\mathbf{1.51 \pm 0.30}$ & \multicolumn{2}{c|}{\cellcolor{red!25}$\mathbf{1.35 \pm 0.34}$} & \multicolumn{2}{c}{-} \\
\cellcolor{blue!25}\textbf{$6$-landmarks Var. Dropout (Dense)} & $1.16 \pm 0.25$ & \cellcolor{blue!25}$\mathbf{0.74 \pm 0.22}$ & $1.60 \pm 0.29$ &  \multicolumn{2}{c|}{$1.54 \pm 0.31$} & \multicolumn{2}{c}{-} \\
\cellcolor{yellow!25}\textbf{$6$-landmarks Targeted Dropout (Dense)} & \cellcolor{yellow!25}$\mathbf{0.76 \pm 0.22}$ & $1.61 \pm 0.28$ & \cellcolor{yellow!25}$\mathbf{1.51 \pm 0.30}$ & \multicolumn{2}{c|}{$1.46 \pm 0.36$} & \multicolumn{2}{c}{-} \\
\textbf{\thead{$9$-landmarks (Dense)}} &  $3.06 \pm 0.37$ & $3.05 \pm 0.37$ & $2.82 \pm 0.35$ & \multicolumn{2}{c|}{$2.42 \pm 0.32$} & \multicolumn{2}{c}{$3.02 \pm 0.33$} \\
\textbf{\thead{Torosdagli et al.~\cite{ness}}} & - & - & - & \multicolumn{2}{c|}{-} & \multicolumn{2}{c}{-} \\
\cellcolor{green!25}\textbf{Gupta et al.~\cite{automated_landmarking_challanges_1}} & $1.42$ & $1.73$ & - & \multicolumn{2}{c|}{$2.08$} & \multicolumn{2}{c}{\cellcolor{green!25}$\mathbf{1.17}$} \\
\bottomrule
\end{tabular}}
\end{table*}

Among two different \textit{RU} architectures, DB architecture was evaluated to be more robust and fast to converge as compared to the \textit{MLP} architecture. To be self-complete, we provided the \textit{MLP} experimental configuration performances only for the $5$-landmark experiment (See Table~\ref{table:experimental_results}).

In the first experiment (Table~\ref{table:setip_input_output}-Experiment 1), to have an understanding of the performance of the RRN, we used the landmark grouping sparsely-spaced and closely-spaced as proposed in Torosdagli et al.~\cite{ness}. We named our first configuration as ``$5$-landmarks'' where closely-spaced, maxillary and nasal bones landmarks are predicted based on the relation of sparsely-spaced landmarks (Table~\ref{table:setip_input_output}). In the $5$-landmarks RRN architecture, there were totally 25 \textit{RUs}. In the second experiment (Table~\ref{table:setip_input_output}-Experiment 2), we explored the impact of a configuration with less number of input mandibular landmarks on the learning performance. Compared to the $5$ sparsely-spaced input landmarks, we learned the relation of $3$ landmarks, $Me$, $Cd_L$ and $Cd_R$, and predicted the closely-spaced landmark locations (as in the $5$-landmarks experiment) plus superior-anterior landmarks $Cor_L$ and $Cor_R$ and maxillary and nasal bones' landmark locations. The network was composed of $9$ \textit{RUs}. The training was relatively fast compared to the $5$-landmarks configuration due to small number of \textit{RUs}. We named this method as ``3-Landmarks Regular''.  

After observing statistically similar accuracy compared to the $5$-landmarks method for the closely-spaced landmarks ($P > 0.005$), and high error rates at the superior-anterior landmarks $Cor_L$ and $Cor_R$, we setup a new experiment which we named ``3-Landmarks Cross'' (Table~\ref{table:setip_input_output}-Experiment 3). In this configuration, the third experiment, we used $1$ superior-posterior and $1$ superior-anterior landmarks on the right and left sides, respectively. This network was similar to $3$-landmarks regular one in terms of number of \textit{RUs} used.

In the fourth experiment (Table~\ref{table:setip_input_output}-Experiment 4), we evaluated the performance of the system in learning the closely-spaced mandibular landmarks $(Gn, Pg, B, Id)$ and the maxillary landmarks $(ANS, A, Pr, PNS)$ using the relation information of the sparsely-spaced and the nasal-bones landmarks which is named as ``$6$-landmarks''. There are a total of $36$ \textit{RUs} in this configuration.

In the last experiment (Table~\ref{table:setip_input_output}-Experiment 5), we aimed to learn the maxillary landmarks $(ANS, A, Pr, PNS)$ and nasal bones landmark $(Na)$ using the relation of the mandibular networks; hence, this network configuration is called ``$9$-landmarks''. The architecture was composed of $81$ \textit{RUs}. Owing to the high number of \textit{RUs} in the architecture, the training of this network was the slowest among all the experiments performed. 

For three challenging CBCT scans, Figure~\ref{fig:results} presents the ground-truth and the predicted landmarks with respect to the $5$-landmarks configuration \textit{DB} architecture, annotated in blue and pink, respectively. We evaluated $5$-landmarks configuration for both \textit{MLP} and the \textit{DB} architectures using variational-dropout as regularizer (Table~\ref{table:experimental_results}). For 4-folds, we observed that \textit{DB} architecture was robust and fast-to-converge. Although, the performances were statistically similar for the mandibular landmarks, this was not the case for the maxillary and the nasal bone landmarks. The performance of the \textit{MLP} architecture degrades notably compared to the decrease in the \textit{DB} architecture for the maxilla and nasal bone landmarks.

$3$-landmarks and $5$-landmarks configurations (Table~\ref{table:experimental_results}) performed statistically similar for the mandibular landmarks. Interestingly, both $3$-landmarks configurations performed slightly better for the neighbouring bone landmarks. This reveals the importance of optimum number of landmarks in the configuration. 

In comparison of $5$-landmarks and $6$-landmarks configurations (Table~\ref{table:experimental_results}), we observed that $5$-landmarks configuration is good at capturing the relations on the same bone. In contrast, $6$-landmarks configuration was good at capturing the relations on the neighbouring bones. Although, the error rates were less than $2mm$, potentially redundant information induced by the $Na$ landmark in the $6$-landmarks configuration caused the performance to decrease notably for the mandibular landmarks compared to the $5$-landmarks configuration.

$9$-landmarks configuration performed statistically similar to $5$-landmarks configuration, however, due to $81$ \textit{RUs} employed for the $9$-landmarks, the training was slower.

Although direct comparison was not possible, we compared our results with Gupta et al.~\cite{automated_landmarking_challanges_1} based on the landmark distances. We found that our results were significantly better for all landmarks except the $Na$ landmark. 
The framework proposed at~\cite{automated_landmarking_challanges_1} uses an initial seed point using a 3D template registration at the inferior-anterior region where fractures are the most common. Eventually, any anatomical deformity that alters the anterior mandible may cause an error in the seed localization which can lead to a sub-optimal outcome.

We evaluated the performance of the proposed system when variational~\cite{Kingma:2015:VDL:2969442.2969527} and  targeted~\cite{gomez2019learning} dropouts were employed. Although there was no accuracy-wise statistically significant difference between dropouts, convergence of the systems were relatively fast (around $20$ epochs compared to $100$ when using regular dropout) for the \textit{MLP} architecture. Hence, for the \textit{MLP} architecture, in terms of computational resources, variational and targeted dropout implementations were far more efficient for our proposed system. This is particularly important because when there are a large number of \textit{RUs}, one may focus more on the efficiency rather than accuracy. When the \textit{DB} architecture was employed, we did not observe any performance improvement among different dropout implementations.

\section{Discussions and Conclusion}
\label{section_4}

We proposed the \textit{RRN} framework which learns spatial dependencies between CMF landmarks in an end-to-end manner. Without the need for an explicit segmentation, we hypothesized that there is an inherent geometrical relation among CMF landmarks which can be learned using the relational reasoning architecture. Although, appearance-based deep-learning approaches are very strong alternative to what we proposed herein, generalization is still an unsolved and a very challenging problem, and reasoning is not directly applicable unlike geometric relations. For instance, authors in~\cite{zheng20153d} used a two-step neural networks with head neck CT data, achieving an average of 2.64 mm localization error; however, their data set does not include any severe pathology, and still performance is inferior to what we have proposed here. The presented solution was shown to be effective in 2D images with normal anatomy. Further, appearance based methods for landmark detection in CT scans \cite{yang2017automatic,ghesu2017robust}, which can be considered related to our work, define landmarks as an anatomical region (ROI) comparatively larger than our landmark definition (25 x 25 vs 3 x 3), and again no deformation or pathology presence exist therein. In contrast to these methods, our method considers a very small area as landmark, and we use extremely challenging pathological cases, which also differentiates the current work from our previous work where we used a segmentation-based approach in the geodesic space. 

Our relational reasoning framework, which is a model-based approach, can generalize well to the unseen data. Hence, once trained, \textit{RRN} can be used at the same testing precision to detect the missing landmarks of the unseen data taken at completely different conditions. This would afford better outcomes for precision medicine and complex CMF deformities. In our experiments, we first evaluated this claim using a dataset with a high amount of bone deformities in addition to other CBCT challenges.  We observed that (1) despite the large amount of deformities that may exist in the CMF anatomy, there is a functional relation between the CMF landmarks, and (2) \textit{RNN} frameworks are strong enough to reveal this latent relation information. Next, we evaluated the detection performance of five different configurations of the input landmarks to find out the optimum configuration. We observed that not all landmarks are equally informative in the detection performance. Some landmark configurations are good in capturing the local information, while some have both good local and global prediction performance. Overall, per-landmark error for the $6$-landmarks configuration is less than $2mm$, which is considered as a clinically acceptable level of success.

In our implementation, we showed that other deep-learning networks can be integrated well into our platform as long as features are encoded via \textit{RUs}. 
While one may argue whether changing specific parameters could make these predictions better. However, such incremental explorations are kept outside the main paper but worth exploring in future studies from an optimization point of view. Moreover, for now RRN only employ spatial information (proof-of-concept stage), its extension could include using shape space learned landmark relationships as a conditional shape prior. Similarly, the use of those learned relationship as a look up table (atlas) is another venue that needs further exploration.

Our study has a number of limitations. For instance, we confined ourselves to manifold data (position of the landmarks and their geometric relations) without use of appearance information because one of our aims was to avoid explicit segmentation to be able to use simple geometric reasoning networks. As an extension of this study, we will incorporate appearance features from medical images to explore whether these features are superior to purely geometric features, or combined (hybrid) features can have additive value in this research domain. One alternative way to pursue the research that we initiated herein will be to explore deeper and more efficient networks. Hence explore how to scale up in to a much wider platform where large number of landmarks and various clinical problems are addressed. We believe that such advances will improve the current technology for 3D visualization and even afford embedding augmented reality to treatment and surgical planning. 

\subsection{Disclosures}
No Conflict of Interest.

\subsection{Acknowledgments}
We thank Mary McIntosh for helping data collection and landmarking. This work is partially supported by the NIH grants: R01-CA246704 and R01-CA240639.

\subsection{Data, Materials, and Code Availability}
Data was collected under IRB approved (PI: Janice S. Lee), and maybe accessible under certain agreement with the NIDCR. Code is available upon request.


\bibliography{relational_reasoning}   
\bibliographystyle{spiejour}   


\vspace{1ex}
\noindent Biographies and photographs of the other authors are not available.

\listoffigures
\listoftables

\end{spacing}
\end{document}